\documentclass[letterpaper, 10 pt, conference]{ieeeconf}  % Comment this line out if you need a4paper

\IEEEoverridecommandlockouts                              % This command is only needed if 
\let\labelindent\relax
\usepackage{mathrsfs}
\usepackage{cite}
\usepackage[colorlinks]{hyperref}
\hypersetup{
    colorlinks=true,
    linkcolor=black,
    citecolor=black,
    filecolor=black,
    urlcolor=black
}
\usepackage{enumitem}
\usepackage{graphics}
\usepackage{epsfig}
\usepackage{booktabs}
\usepackage{ragged2e}
\usepackage{multirow}
\usepackage{array}
\usepackage{mathrsfs}
\usepackage{amsmath}
\usepackage{amsfonts}
\usepackage{amssymb}
\usepackage{makecell}
\usepackage{algorithm}
\usepackage{siunitx}
\usepackage{url}
\usepackage{caption}
\title{\LARGE \bf
Learning Instruction-Guided Manipulation Affordance via Large Models for Embodied Robotic Tasks
}
\author{Dayou Li\textsuperscript{\&}, Chenkun Zhao\textsuperscript{\&}, Shuo Yang, Lin Ma, Yibin Li, and Wei Zhang${^*}$% <-this % 
\thanks{Dayou Li, Chenkun Zhao, Shuo Yang, Yibin Li, and Wei Zhang are with Shandong University, China. Lin Ma is with Meituan.}
\thanks{\textsuperscript{\&}These authors contributed equally to this work.}
\thanks{${^*}$Corresponding author: Wei Zhang (email: davidzhang@sdu.edu.cn)}
\thanks{This work was supported in part by the National Natural Science Foundation of China under Grants 61991411 and U22A2057, and in part by the Project for Self-Developed Innovation Team of Jinan City under Grant 2021GXRC038.}}

\begin{document}

\maketitle
\thispagestyle{empty}
\pagestyle{empty}

%%%%%%%%%%%%%%%%%%%%%%%%%%%%%%%%%%%%%%%%%%%%%%%%%%%%%%%%%%%%%%%%%%%%%%%%%%%%%%%%
\begin{abstract}
We study the task of language instruction-guided robotic manipulation, in which an embodied robot is supposed to manipulate the target objects based on the language instructions. In previous studies, the predicted manipulation regions of the target object typically do not change with specification from the language instructions, which means that the language perception and manipulation prediction are separate. However, in human behavioral patterns, the manipulation regions of the same object will change for different language instructions. In this paper, we propose Instruction-Guided Affordance Net (IGANet) for predicting affordance maps of instruction-guided robotic manipulation tasks by utilizing powerful priors from vision and language encoders pre-trained on large-scale datasets. We develop a Vison-Language-Models(VLMs)-based data augmentation pipeline, which can generate a large amount of data automatically for model training. Besides, with the help of Large-Language-Models(LLMs), actions can be effectively executed to finish the tasks defined by instructions. A series of real-world experiments revealed that our method can achieve better performance with generated data. Moreover, our model can generalize better to scenarios with unseen objects and language instructions.
\end{abstract}

%%%%%%%%%%%%%%%%%%%%%%%%%%%%%%%%%%%%%%%%%%%%%%%%%%%%%%%%%%%%%%%%%%%%%%%%%%%%%%%%
\section{Introduction}
\begin{figure}[t]
\begin{center}
\includegraphics[width=1\linewidth]{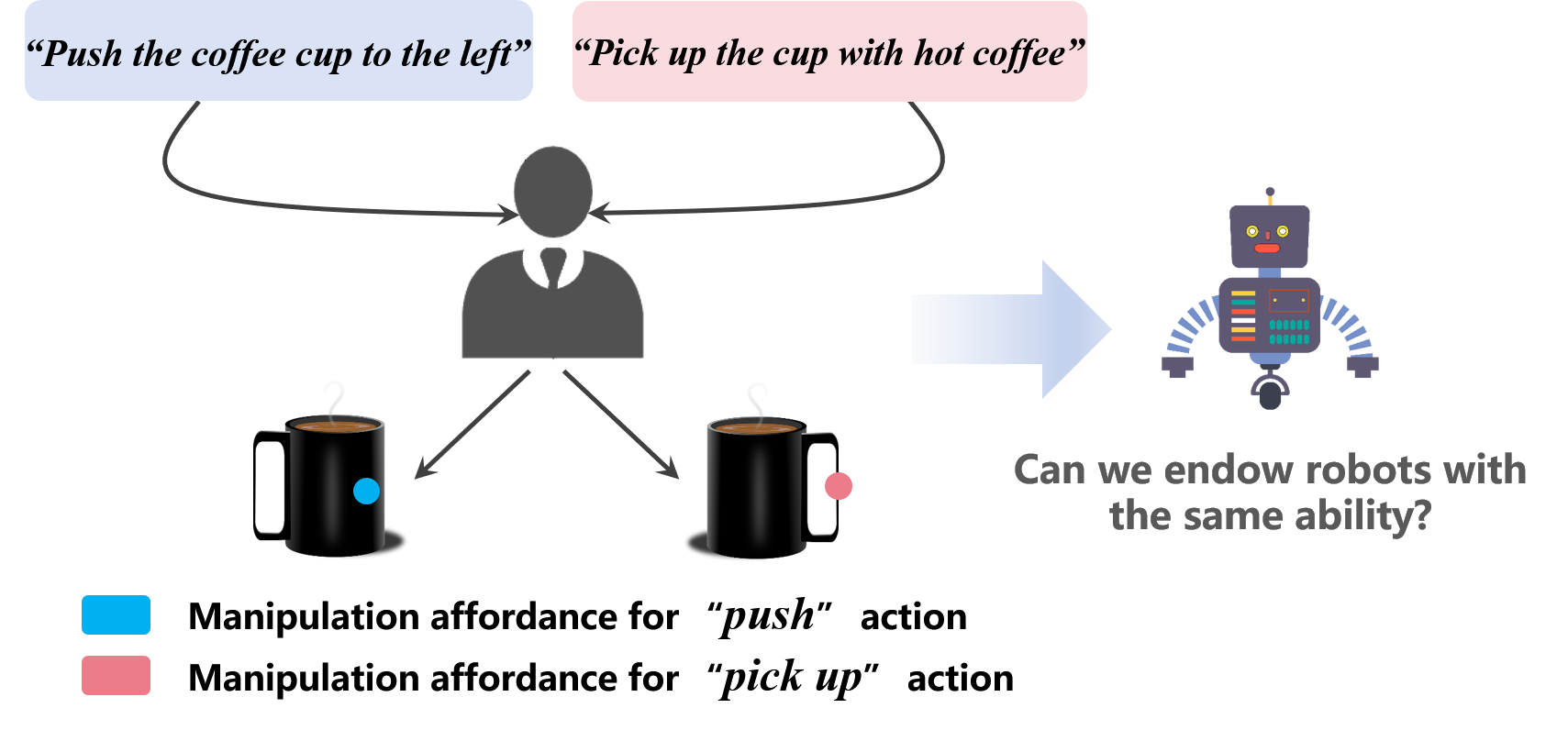}
\end{center}
\caption{Illustration of human's reasoning process in handling instruction-guided manipulation tasks.}
\label{introd}
\end{figure}

% 加语言指令，一是指定操作物品，二是指定操作方式
% 先介绍几篇重点在于操作物品的
Humans are able to perform diverse tasks according to language instructions. Embodied robots are expected to possess ``human-like'' manipulation abilities in daily lives. However, it is non-trivial to endow robots with the same understanding ability and manipulation flexibility as humans. Fig.~\ref{introd} shows a typical manipulation example. If you ask a person to `\textit{push the coffee cup to left}', they will manipulate the main body of the coffee cup. But when the instruction turns to `\textit{pick up the cup with hot coffee}', they tend to grab the handle of the coffee cup due to its hot temperature. This example shows that humans can select appropriate parts of an object for manipulation based on different instructions. Can we endow robots with the same capability of selecting appropriate parts of objects for instruction-guided manipulation?

% Ask a person to `push the coffee cup to me' and they can naturally take the concept of the mug and ground it in real-world action. By human intuitions, they will manipulate the main body of the mug according to the instruction 'push'. But when you ask a person to 'pass me the hot coffee', they will grab the handle of the mug and carry it to your front instead of pushing it due to its hot temperature.

Manipulation affordance indicates functional interactions of object parts with humans, which is considered an effective manipulation-centric representation for enabling diverse embodied robotic tasks. In fact, it is not a new thing to learn object manipulation affordance \cite{10064325,9363610}. Zeng et al. \cite{zeng2020tossingbot,zeng2018learning} propose to learn action affordance by self-supervised learning or supervised learning. The training data is obtained by agent exploring in simulation or human labeling. Lin et al. \cite{yen2020learning} attempt to directly transfer model parameters from vision models to affordance prediction networks. Mo et al. propose Where2Act \cite{mo2021where2act} to predict the action affordance of 3D objects. Where2Act is capable of selecting suitable manipulation parts according to action primitives while still lacking the ability to handle language instructions. Recent progress in large language models and multi-modal models makes it easier to incorporate language instructions for manipulation affordance prediction. CLIPORT \cite{shridhar2022cliport} integrates semantic understanding of CLIP \cite{radford2021learning} and spatial precision of Transporter \cite{zeng2021transporter} together into a convolutional architecture to predict manipulation affordance based on language instructions. Luo et al. \cite{luo2023grounding} present a two-stage framework for grounding spatial-related instructions for affordance predictions in object manipulation tasks, which also apply CLIP to extract features of language and image input. The above methods are built upon the pre-trained CLIP model for feature extraction and then train a convolutional architecture to predict affordance maps. However, they collect data in a simulation environment or human demonstration, causing a heavy human burden and a sim2real gap.
% Besides, the CLIP features can also be used as labels to train neural networks. For example, Shafiullah et al. \cite{shafiullah2022clip} trained real-world CLIP-Fields using CLIP visual representation as supervision.

Recently, Large Language Models (LLMs) have been employed in robotic manipulation tasks and achieved significant progress \cite{lin2023text2motion,song2023llm,ding2023task,ahn2022can}. In particular, Ahn et al. \cite{ahn2022can} leveraged the probability distribution of text output of GPT-3 by calling API and combined it with value function to generate affordance value of the actions that will be probably performed in the next state. Unfortunately, this method currently only supports affordance prediction of task level. 
% Xu et al. \cite{xu2023joint} use CLIP for image and text feature extraction, as well as pre-trained GraspNet \cite{fang2020graspnet} to provide grasp feature query, and then jointly modeled them by a cross-attention transformer to predict grasp logits and values. However, these methods are not able to predict instruction-guided manipulation affordance for facilitating diverse embodied robotic tasks.
% 说一下我们的framework的有点，可以在像素级别做affordance预测，也可以对任务进行分解
% 提一下我们自己标注了一些数据集，这是为了使扩充的数据更具有针对性

\begin{figure*}[t]
\begin{center}
\includegraphics[width=1\linewidth]{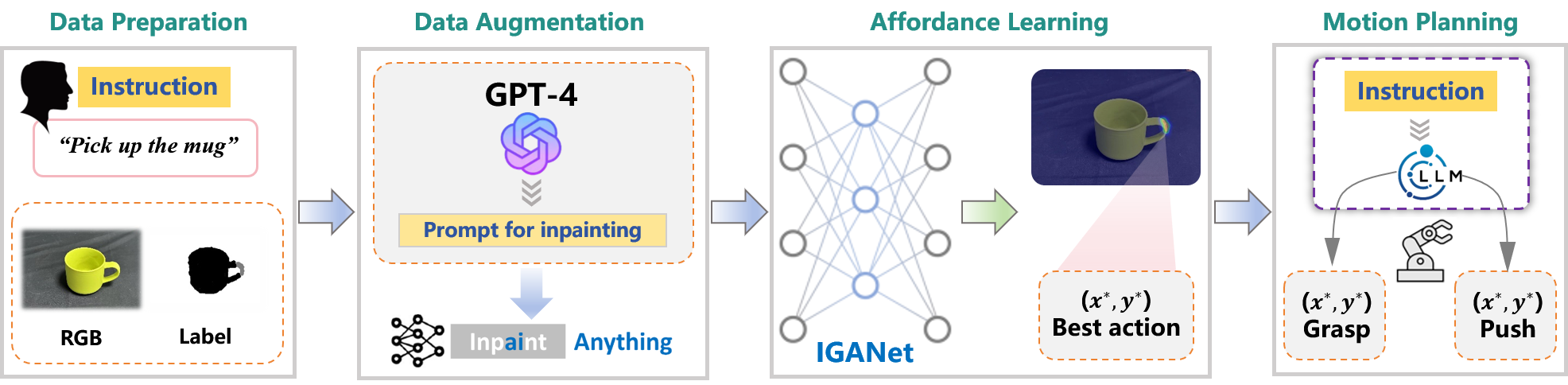}
\end{center}
\caption{Illustration of the presented full pipeline for instruction-guided manipulation tasks. Our pre-labeled dataset will be scaled up via our data augmentation pipeline. Then the IGANet is trained on the generated dataset to predict affordance maps based on the language instructions. Finally, the LLM-based planner will give commands on action execution based on the affordance maps and instructions.}
\label{fullpipe}
\end{figure*}

In this paper, we aim to propose an efficient pipeline for instruction-guided robotic manipulation tasks. To this end, three core components are presented, including a manipulation affordance prediction model of pixel level, a data augmentation pipeline via Vision-Language Models (VLMs), and an automatic action planner. Fig.~\ref{fullpipe} illustrates the full pipeline that integrates the three components. Specifically, we design a data augmentation pipeline based on Inpaint-Anything \cite{yu2023inpaint}, which can edit the image based on our instruction generated by GPT-4. We manually label a small amount of real-world data and then expand the dataset through our proposed pipeline to ensure that the generated data will be similar to the real-world data. We apply a pre-trained OWL-ViT vision encoder and Universal-Sentence-Encoder to produce vision and language features for affordance prediction. Besides, an LLM-based action planner is present to determine action according to the language instructions, ensuring that the instructions can be well completed. 
To summarize, our main contributions are:
\begin{itemize}[leftmargin=*]
    \item We propose IGANet, an efficient framework for learning instruction-guided robotic manipulation affordance, which jointly models language and vision. An LLM-based action planner is also proposed for action planning to guide the robot through abstract instructions.
    \item We present an automatic data augmentation pipeline using a diffusion model as an VLM and an LLM, which can produce a large amount of data for model training.
    \item The proposed method is evaluated on a series of scenarios with seen and unseen objects and language instructions, demonstrating the effectiveness and generalization of our framework.
\end{itemize}

\section{Related Works}

\subsection{VLMs-Driven Data Augmentation}
% With the increasing sophistication of large diffusion model for text-image generation, many researchers have applied diffusion models for dataset expansion \cite{brooks2023instructpix2pix,zeng2023distilling,yu2023scaling}.
% Mainstream image generation diffusion models are capable of generating a completely new image or editing the input image based on the input language instructions. Also, LLMs are able to generate a number of instructions according to the user's preference. Thus, we can build a automatic data augmentation pipeline using VLMs and LLMs to generate data as large-scale as we want.

Nowadays, Vision-Language models have shown strong capabilities in understanding image-text pairs and image generation. DALL-E, powered by OpenAI, as well as StableDiffusion\cite{rombach2021highresolution} and Midjourney, are powerful image generators that can generate an image based on the language description. They are trained on web-scale datasets and thus have strong generalization. Many scholars apply such generative models for dataset expansion \cite{brooks2023instructpix2pix,zeng2023distilling}. Zeng et al. \cite{zeng2023distilling} apply VLM to generate goal rearrangement images based on the structured scene descriptions. Brooks et al. \cite{brooks2023instructpix2pix} adopt GPT-3 to generate image editing instructions and use StableDiffusion\cite{rombach2021highresolution} to generate edited images. Yu et al. \cite{yu2023scaling} adopt DALL-E to scale up data for robotic manipulation task learning. Access to such generative VLMs enables us to automate the generation of large-scale datasets. 

With such inspiration, we now review image editing techniques that can be used in our data augmentation pipeline. Image editing targets pixel-level editing of images for tasks such as style transfer, background replacement, image insertion, object removal, etc\cite{gatys2016image,hertz2022prompt,laput2013pixeltone,meng2021sdedit,kawar2023imagic}. From Generative Adversarial Networks (GANs) to Diffusion Models, image editing methods have also been revolutionized. The sophisticated GAN-based methodology is gradually losing its dominance in image editing topic. When equipped with large-scale image-text datasets, such as Laion-5b\cite{schuhmann2022laion} and InstructPix2Pix \cite{brooks2023instructpix2pix}, diffusion models are endowed with powerful generative capabilities. Many image editing models such as Object 3dit\cite{michel2023object} and InstructPix2Pix\cite{brooks2023instructpix2pix} have shown stable ability to edit image based on an input image and a text instruction of how to edit it. Therefore, we intend to apply this technique to expand our human-labeled small-batch data for robotic manipulation affordance.

\subsection{Language-Guided Robotic Manipulation}
Because of the availability of Large-Language Models (LLMs) and Vision-Language Models (VLMs), language-guided robotic manipulation has become a spotlight research topic in recent years. Chen et al.\cite{chen2021joint} focus on the task of grasping the target object based on a natural language command query. They adopt LSTM as a language command encoder. However, after Radford et al. released CLIP\cite{radford2021learning}, the way in which language commands and images are encoded has also changed dramatically. Many scholars attempt to apply text encoder and image encoder in CLIP for their model, as CLIP is trained by a web-scale dataset. Shridhar et al. \cite{shridhar2022cliport} present CLIPORT, which is a language-conditioned imitation learning agent that combines the broad semantic understanding of what CLIP\cite{radford2021learning} with spatial precision. Xu et al. \cite{xu2023joint} propose to jointly model vision, language, and action with object-centric representation by using both image and text encoders in CLIP. Shafiullah et al. \cite{shafiullah2022clip} adopt CLIP embeddings to train real-world CLIP-Fields for goal navigation. Shen et al. \cite{shen2023distilled} extract dense features from CLIP using the MaskCLIP reparameterization trick to support zero-shot language guidance.
Rather than applying CLIP for feature extraction in language-guided manipulation, some scholars also use LLMs for action planning. Wu et al. \cite{wu2023tidybot} leverage the summarization capabilities of LLMs to infer generalized user preferences by planning which object should be manipulated. Sharan et al. \cite{sharan2024plan} employ LLMs to generate single-step text sub-goals that will be translated into vision sub-goals via a diffusion model.

In conclusion, most pipelines that adopt LLMs as planners can only perform coarse planning for tasks. However, tasks such as fine-grained planning for object manipulation regions, usually require supervise-learning methods. However supervise-learning algorithms are in need of large-scale datasets, which are always difficult to obtain. Thus, we aim to combine the generative power of VLMs with the strong feature extraction capability of CLIP and the excellent planning capability of LLMs, in order to overcome the shortcomings respectively. 

\begin{figure}[t]
\begin{center}
\includegraphics[width=1\linewidth]{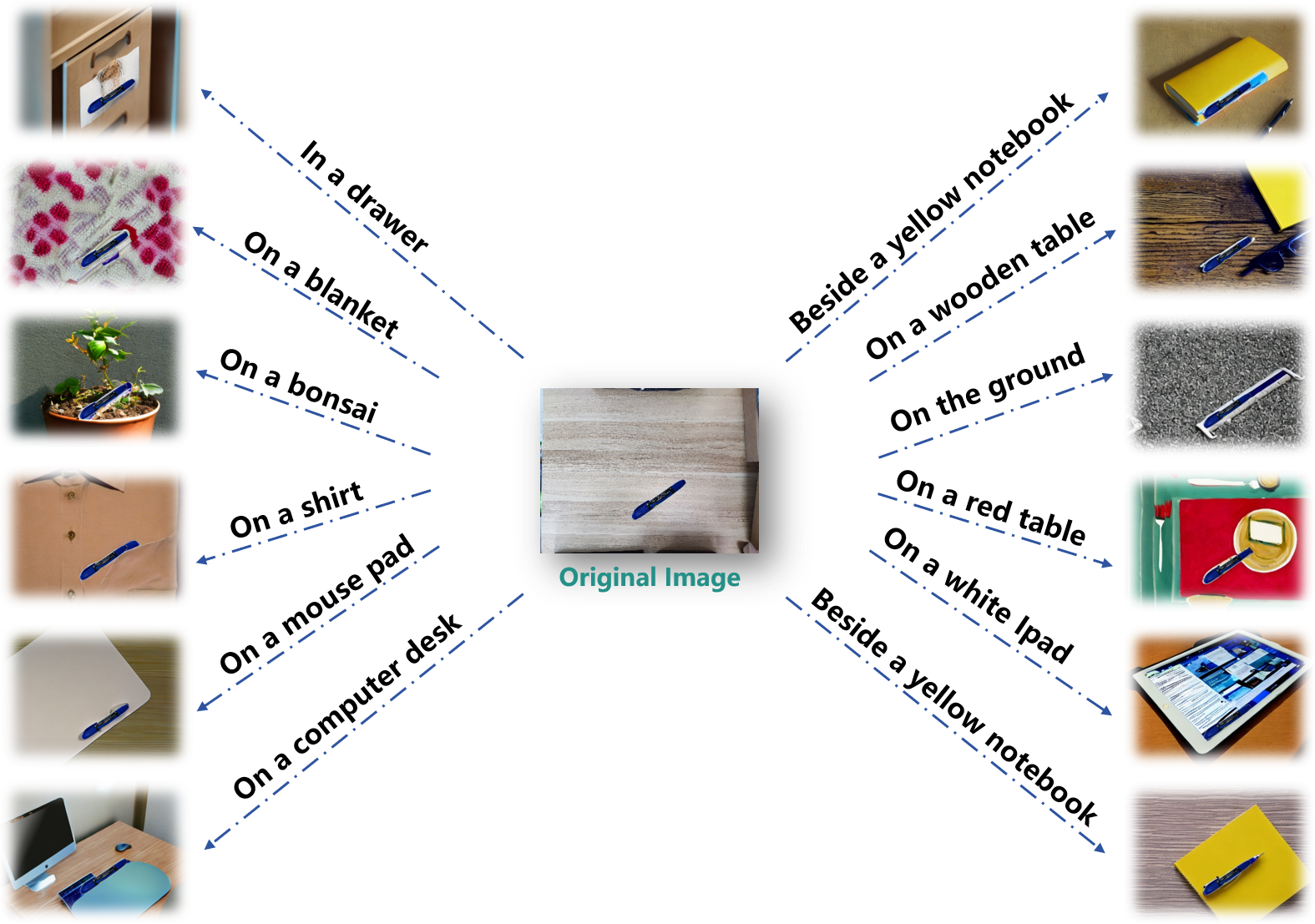}
\end{center}
\caption{Data Generation Result. Our proposed data augmentation pipeline uses GPT-4 as an LLM to generate prompts for the Inpaint-Anything module to edit the image according to the generated prompts.}
\label{datagen}
\end{figure}

\section{Method}

\subsection{Pipeline Overview}
As shown in Fig.~\ref{fullpipe}, we present Instruction-Guided Affordance Net (IGANet), a novel method for learning instruction-guided manipulation affordance. Given a prompt that defines the manipulation instruction and an image $I \epsilon$ $\mathbb{R}^{H\times W \times 3}$ of the target object, our objective is to produce object manipulation affordance based on the task instruction. Particularly, IGANet takes as input a text prompt denoted by $p$, demonstrating which object is set as the target and how it will be manipulated. The desired output of our network is an affordance map $M$ that assigns an affordance value to each pixel of the input image, representing the operable area of the target object for the task description. Besides, we propose a data augmentation pipeline using Vision-Language Models (VLMs), which is capable of generating diverse desired data automatically. Finally, with our proposed action planner, the language instructions can be broken down into executable actions, which our real-world robotic platform can perform.
% 配一张系统图

\subsection{Scaling up Data via VLMs} 
% In our setup, the maneuverable area of the same object is different for different tasks. However, there are no publicly available dataset for such tasks, and our manually labeled dataset is insufficient for good generalization performance of the model. We propose to apply VLMs to expand our dataset. So far, VLMs are not capable of auto-labeling manipulation affordance that we label in our proposed dataset. There are some VLMs that have excellent ability of image-editing \cite{zhang2023diffusionengine,rombach2022high,brooks2023instructpix2pix,wang2023imagen,yu2023inpaint}. Therefore, we propose to apply the image-inpainting technique in our pipeline to augment our data automatically. 
% 预先标注一些物品的affordance，定义任务；让GPT根据任务编写inpainting的prompt，这里的prompt跟任务无关，只是调整图像变化
% Inpainting-Anything根据prompt进行图片编辑
First, for a given object, we label different manipulation affordance according to different tasks. Then, we apply LLMs to generate some inpainting prompts that define the changes the model should make to the images. Inpaint-Anything \cite{yu2023inpaint} merges SAM\cite{kirillov2023segment}, LaMA\cite{suvorov2022resolution}, and StableDiffusion\cite{rombach2022high} to enable the user to remove, fill and replace anything in the image. Inspired by Inpaint-Anything, we propose our pipeline to augment our data.
% 配图

We pre-define 30 items and label their manipulation affordance with language instructions according to the task, along with their masks. We rotate the objects, and their masks, and labeled affordance in the original image to enlarge our initial dataset. Even though a hand-engineered prompt may guarantee the generated data to be out-of-distribution, the size of the generated prompts is not large enough. Therefore, we leverage common sense learned by LLMs to provide various prompts for image editing with detailed descriptions of the imagined scene. By processing the generated prompts, Inpaint-Anything can edit the image into various styles, as shown in Fig.~\ref{datagen}. Also, we query the LLM to generate diverse language instructions for manipulation, which share similar meanings to our human-labeled data but are expressed in a different form. Using GPT-4 to generate rather than manually formulate prompts ensures their diversity. To sum up, the data format that we generate is text-image-affordance.

\begin{figure}[t]
\begin{center}
\includegraphics[width=1\linewidth]{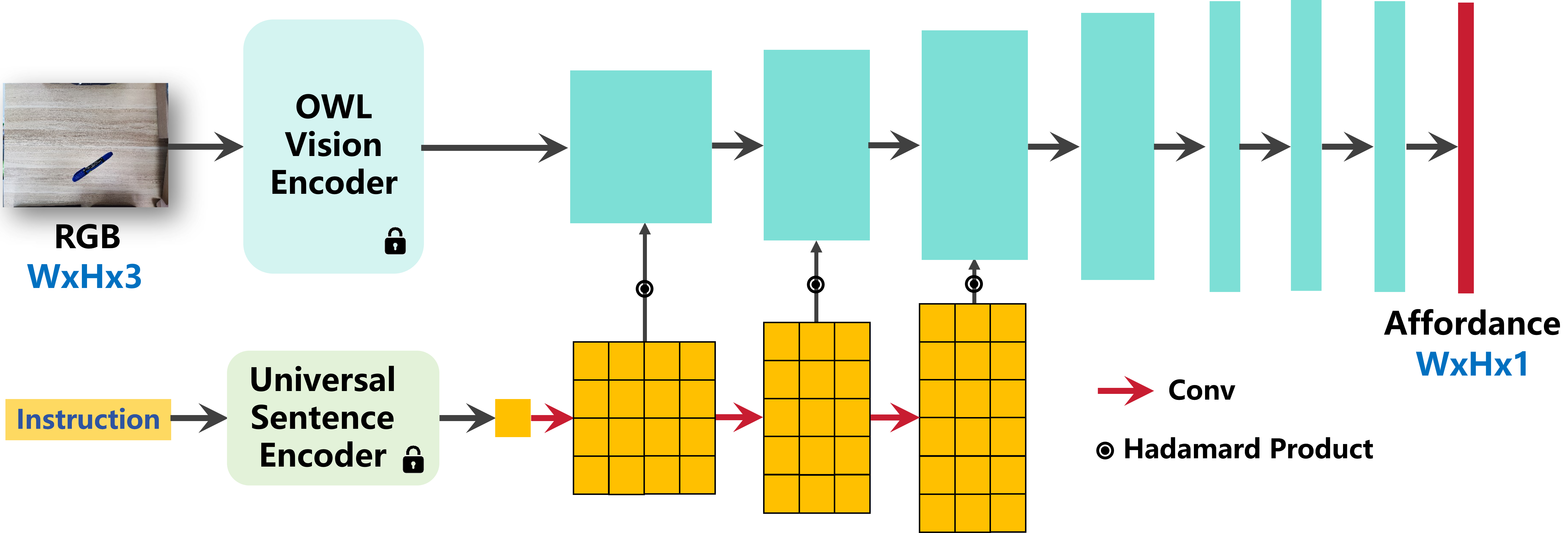}
\end{center}    
\caption{Structure of IGANet. IGANnet uses a frozen OWL vision encoder to encode RGB input, and the language instruction is encoded by a frozen Universal-Sentence encoder. The RGB feature and language feature perform Hadamard Product operation. The final output of IGANet is the affordance map of dense pixel-wise features.}
\label{networkdetail}
\end{figure}

\subsection{Learning Instruction-Guided Manipulation Affordance}

In IGANet, as shown in Fig.~\ref{networkdetail}, we propose to yield visual features through OWL-ViT \cite{minderer2205simple} vision encoder denoted by $E_{vit}$ and language features through Universal-Sentence-Encoder \cite{cer2018universal} denoted by $E_{uni}$. The frozen OWL-ViT vision encoder encodes RGB input to produce dense features. Then we propose a decoder consisting of several fully-connected convolutional layers and some upsample layers in between. 

The instruction prompt is first encoded by the text encoder $E_{uni}$ to produce a goal encoding $g=E_{uni}(p)$. The goal encoding $g$ is then downsampled with a fully-connected layer to keep consistent with the channel dimension. Then the goal encoding is tiled to match with the spatial dimension of the decoder features. We take the Hadamard product of the decoder features and the tiled goal encodings, this element-wise product enables us to combine instruction prompt features with image features, achieving alignment between both. This instruction guiding is conducted repeatedly for three consecutive layers of the decoder. Finally, the channel dimension to 1 is reduced by applying $1\time1$ convolutional layer to the decoded features, producing manipulation affordance with ReLU function. 

During training, we apply cross-entropy loss to train our model:
\begin{equation}
    L=-\sum_{(r, c)}^{(w, h)}[P_{G(r,c)}\log P_{A(r, c)}]
\end{equation}
where $(r,c)$ denotes the pixel coordinate and $(w,h)$ denotes the shape of the image. $P_{G(r,c)}$ and $P_{A(r, c)}$ represent ground truth and predicted affordance respectively.

\begin{figure}[t]
\begin{center}
\includegraphics[width=1\linewidth]{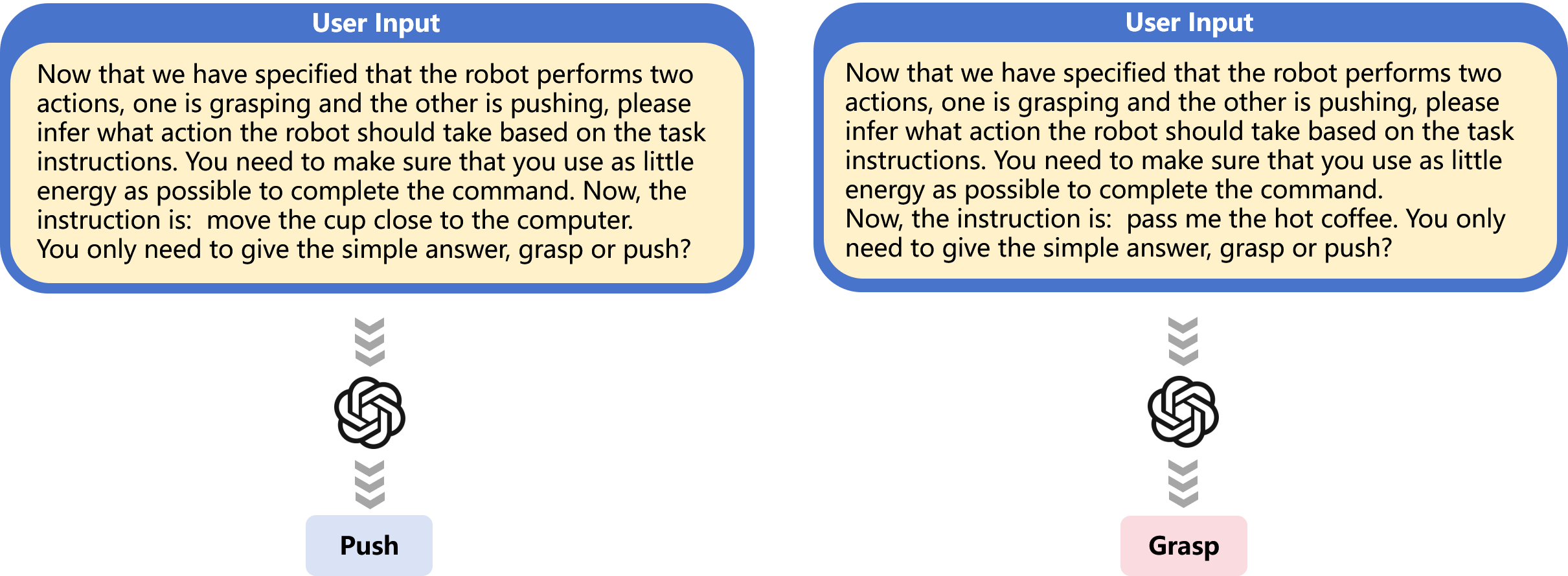}
\end{center}    
\caption{LLM-Based Planner. The LLM-based planner uses GPT-4 as LLM to give action decisions based on our prompt engineering.}
\label{action planner}
\end{figure}

\subsection{Action Execution}
We define two primitive actions including grasping and pushing, the following contains some specific parameters of the actions and how we propose to determine them:
\begin{itemize}[leftmargin=*]
    \item \emph{Grasping}: A grasping action can be defined as $a_{g}=(p_{g},\theta_{g})$, where $p_{g}=(x_{g},y_{g},z_{g}) \in \mathbf{R}^{3}$ represents the middle position of the top-down parallel-jaw grasp and $\theta_{g} \in \mathbf{R}$ represents the grasping angle that ranges within $-90\si{\degree}$ to $90\si{\degree}$ around the z-axis. We use DINO\cite{liu2023grounding} to predict the target bounding box as box prompt for SAM\cite{kirillov2023segment} to acquire segmentation mask of the object. Then we calculate the short side tilt angle of the target's bounding box as the rotation angle. $h_{g}$ is the height at point $(x_{g},y_{g})$, and $z_{g}=h_{g}-2cm$. The gripper needs to move down 2 cm below $h_{g}$ to execute the grasping action.
    \item \emph{Pushing}: A pushing action can be defined as $a_{p}=(p_{p},d_{p})$ that is performed by the tip of the gripper. Each push length is fixed at 13 cm and the pushing trajectory is straight. $p_{p}=(x_{p},y_{p},z_{p}) \in \mathbf{R}^{3}$ represents the starting position of pushing action and $d_{p} \in \mathbf{R}^{3}$ is the pushing direction vector.
    The end position of the pushing action is derived by GPT-4 with vision (GPT-4V) based on the input language instruction. GPT-4V takes as input a description of the scene, including object categories and their bounding boxes, along with the manipulation instructions and the image of the scene. Relying on the excellent detection capability of DINO, the end position of the pushing action can be roughly estimated. Thus the direction vector can be calculated. %Besides, $(x_{p},y_{p})$ is the position calculated from pixel in predicted affordance by calibration. 
\end{itemize}
Moreover, as shown in Fig.~\ref{action planner}, the GPT-4V is also able to make selections of action according to the language manipulation instructions via prompt engineering.

\begin{table*}[t]\small
\centering
\caption{Real-World Experiment Results}\label{tab_res}
% \resizebox{15cm}{!}{
\begin{tabular}{p{4.4cm}m{2.2cm}<{\centering}m{2.2cm}<{\centering}m{2.2cm}<{\centering}m{2.2cm}<{\centering}m{2.2cm}<{\centering}m{2.2cm}<{\centering}}%{ccccccc}
\toprule[1pt]
\multirow{2}{*}{Method} & \multicolumn{6}{c}{Success Rate}                                                                                    \\ \cmidrule{2-7}%\cline{2-7} 
                        & \multicolumn{1}{c}{Scene 1} & \multicolumn{1}{c}{Scene 2} & \multicolumn{1}{c}{Scene 3} & \multicolumn{1}{c}{Scene 4} & \multicolumn{1}{c}{Scene 5} & Scene 6 \\ \midrule[0.5pt]%\hline
GPT + ViT                 & \multicolumn{1}{c}{60\% (12/20)}    & \multicolumn{1}{c}{50\% (10/20)}    & \multicolumn{1}{c}{60\% (12/20)}    & \multicolumn{1}{c}{45\% (9/20)}    & \multicolumn{1}{c}{55\% (11/20)}    & 50\% (10/20)    \\ \midrule[0.5pt]%\hline
GPT + DINO                & \multicolumn{1}{c}{50\% (10/20)}    & \multicolumn{1}{c}{55\% (11/20)}    & \multicolumn{1}{c}{40\% (8/20)}    & \multicolumn{1}{c}{50\% (10/20)}    & \multicolumn{1}{c}{70\% (14/20)}    & 65\% (13/20)    \\ \midrule[0.5pt]%\hline
Ours w/o data augmentation & \multicolumn{1}{c}{70\% (14/20)}    & \multicolumn{1}{c}{75\% (15/20)}    & \multicolumn{1}{c}{85\% (17/20)}    & \multicolumn{1}{c}{25\% (5/20)}    & \multicolumn{1}{c}{35\% (7/20)}    & 30\% (6/20)    \\ \midrule[0.5pt]%\hline
\textbf{Ours}                    & \multicolumn{1}{c}{\textbf{75\% (15/20)}}    & \multicolumn{1}{c}{\textbf{85\% (17/20)}}    & \multicolumn{1}{c}{\textbf{90\% (18/20)}}    & \multicolumn{1}{c}{\textbf{80\% (16/20)}}    & \multicolumn{1}{c}{\textbf{90\% (18/20)}}    & \textbf{95\% (19/20)}    \\
\bottomrule[1pt]
\end{tabular}
\end{table*}

% \textbf{Vision-language Model.}
\begin{figure}[bt]
\begin{center}
\includegraphics[width=1\linewidth]{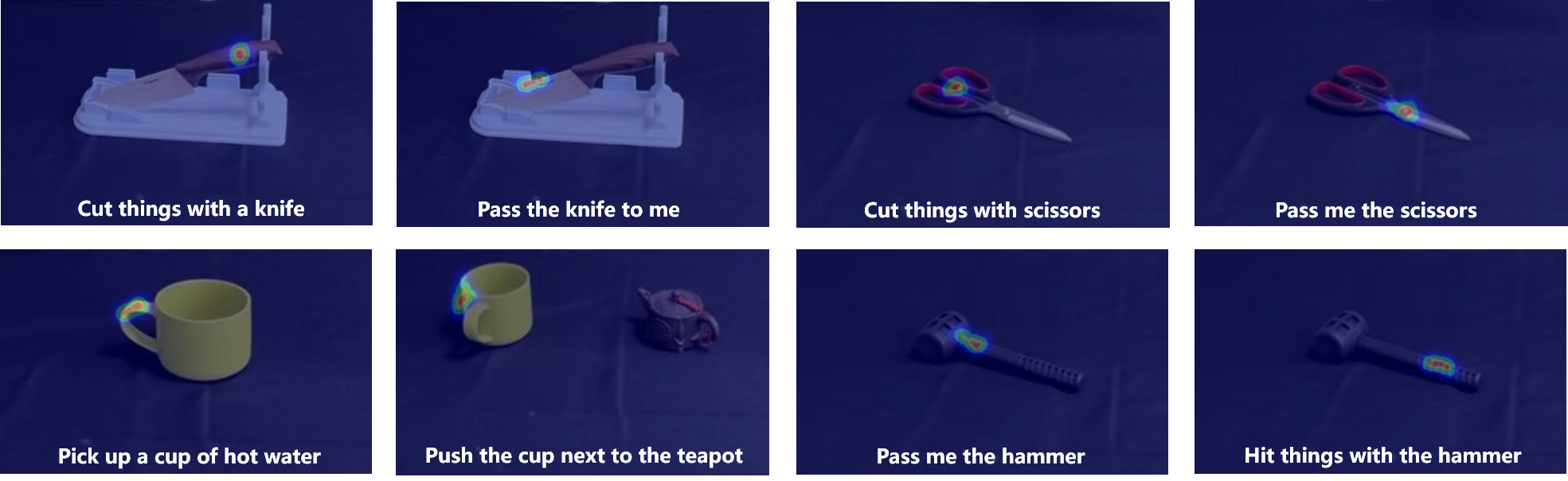}
\end{center}
\caption{Affordance Prediction. Here shows several visualizations of predicted affordance maps based on language instructions.}
\label{affordance result}
\end{figure}

\begin{figure*}[t]
\begin{center}
\includegraphics[width=1\linewidth]{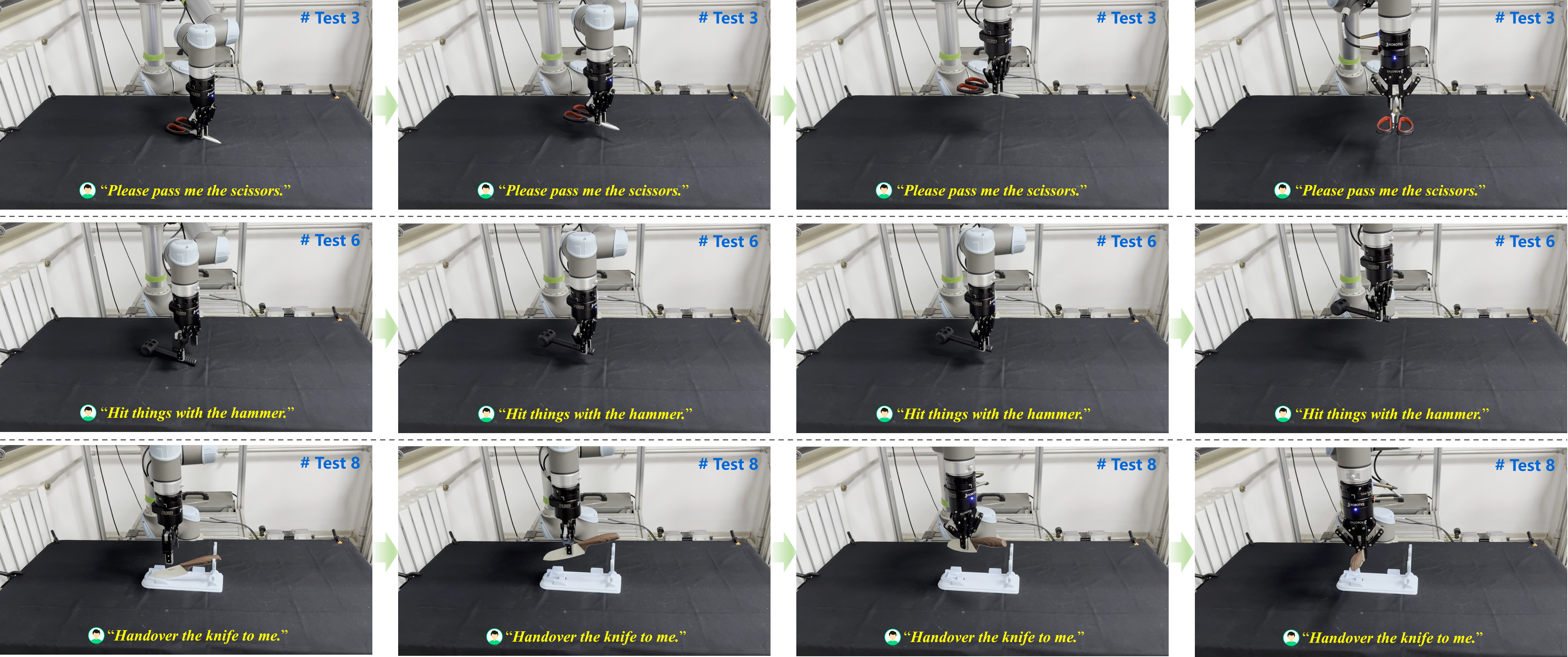}
\end{center}
\caption{Real-world Execution. Examples of the real-world test results performed by our robotic system.}
\label{real execution}
\end{figure*}

\section{Experiments}
We conduct real-world experiments to evaluate our method. The goals of our experiments are: 1) to demonstrate that our VLM-driven data augmentation pipeline is effective in boosting the performance of our proposed model; 2) to evaluate the generalization of our model on unseen objects and language instructions; 3) to validate the effectiveness of our overall framework. Details are as follows.
% 对比方法：
% Distilled Feature Fields Enable Few-Shot Language-Guided Manipulation
% 如果仅用手工标注的数据训练，换场景测试affordance或整体任务；换物体测试；对比一下其他方法（这里想想对比哪个，能够体现指令影响操作区域的）

\subsection{Environment Setup}
% 系统构成，训练集有哪些物体，任务指令都有哪些
To test the real-world performance of our method, we develop an embodied robotic system to perform instruction-guided manipulation tasks. The developed system is composed of a UR5 robotic arm, an Intel RealSense D435i camera that is used to capture the observation of the workspace, and a ROBOTIQ two-finger gripper mounted on the end of the robotic arm. We set up six different scenarios with a series of objects. In each scene, we randomly select several classes of objects and create 20 different testing tasks using different instances of selected classes and diverse language instructions. A test can be judged as a success if the robot successfully manipulates the objects according to the given instructions. We report the task success rate to measure the performance of our method.

\subsection{Baseline Methods}
% We compare the performance of your system with the following baselines:
Given that no works adopt the same technical pipeline as us in the field of instruction-guided manipulation, we create two baseline methods based on currently popular large models. We use GPT-4 as the instruction reasoning and task planning engine as it is one of the most widely used LLMs showing powerful reasoning ability. For visual perception and manipulation frame generation, we employ two popular VLMs, OWL-ViT \cite{minderer2205simple} and Grounding DINO. We term the two baseline methods GPT+ViT and GPT+DINO, respectively. To further validate the proposed VLM-driven data augmentation strategy, we additionally create a variant of our full method by removing the data augmentation, termed as Ours w/o data augmentation.

% \textbf{GPT+CLIP}
% is a variant of \textbf{GPT+ViT}. The difference is that \textbf{GPT+DINO} uses CLIP\cite{radford2021learning} as an object detector for generating operation points.

% \textbf{Ours without data augmentation}
% is an implementation of our proposed framework without generated data for training. Thus, the training data is only the human-label part.

\subsection{Results}
The comparative results are provided in Table.~\ref{tab_res}. The results show that our method outperforms all baseline methods by a large margin across six different scenes, validating the effectiveness of our method in performing instruction-guided manipulation tasks. The first 3 scenes contain seen objects and language instructions, while the remains contain unseen ones. The ``GPT+X"  methods use open-vocabulary detectors for detecting the target object and generating the center point of the bounding box as the operation position. However, from the result, we can find that such methods do not surpass us in performance, which demonstrates that determining the manipulation region by the center of the bounding box is not flexible enough to deal with versatile language instructions. From Table ~\ref{tab_res}, we can see that our method without our VLM-aided data generation performs badly in unseen scenarios. This is because the amount of the human-labeled data is not large enough and the model is overfitting. By training the IGANet with our generated data, we observe that the performance of our entire framework is relatively satisfactory, and our proposed VLM-based data augmentation pipeline can alleviate the loss of model performance caused by a lack of data. We provide the visualization examples of manipulation affordance generated by IGANet in Fig.~\ref{affordance result}. It can be seen that IGANet can accurately predict different-located manipulation affordance on objects conditioned on the given instructions. Fig.~\ref{real execution} shows some testing cases of real-world experiments, demonstrating reliable instruction-guided manipulation performance on a real robot. Full real-world demonstrations of our method can be seen on \url{https://youtu.be/tgQ_K1Yj2c0}.

\section{Conclusions}
In this work, we focus on the task of instruction-guided robotic manipulation. We take full advantage of LLMs and VLMs in our proposed framework. Faced with the situation of a small amount of data, we propose a VLM-based data augmentation pipeline to generate a large amount of data automatically for model training. We apply LLMs in our action planner to assist in action execution, ensuring the actions can be performed effectively. We also utilize the pre-trained vision and language encoder in our proposed affordance prediction model (IGANet). It is worth mentioning that our method incorporates the currently popular large model technique, which provides new ideas for language-guided robotic manipulation tasks.

% \addtolength{\textheight}{-12cm}   % This command serves to balance the column lengths
%                                   % on the last page of the document manually. It shortens
%                                   % the textheight of the last page by a suitable amount.
%                                   % This command does not take effect until the next page
%                                   % so it should come on the page before the last. Make
%                                   % sure that you do not shorten the textheight too much.

%%%%%%%%%%%%%%%%%%%%%%%%%%%%%%%%%%%%%%%%%%%%%%%%%%%%%%%%%%%%%%%%%%%%%%%%%%%%%%%%

%%%%%%%%%%%%%%%%%%%%%%%%%%%%%%%%%%%%%%%%%%%%%%%%%%%%%%%%%%%%%%%%%%%%%%%%%%%%%%%%

%%%%%%%%%%%%%%%%%%%%%%%%%%%%%%%%%%%%%%%%%%%%%%%%%%%%%%%%%%%%%%%%%%%%%%%%%%%%%%%%
% \section*{APPENDIX}

% Appendixes should appear before the acknowledgment.

% \section*{ACKNOWLEDGMENT}

% The preferred spelling of the word ÒacknowledgmentÓ in America is without an ÒeÓ after the ÒgÓ. Avoid the stilted expression, ÒOne of us (R. B. G.) thanks . . .Ó  Instead, try ÒR. B. G. thanksÓ. Put sponsor acknowledgments in the unnumbered footnote on the first page.

% %%%%%%%%%%%%%%%%%%%%%%%%%%%%%%%%%%%%%%%%%%%%%%%%%%%%%%%%%%%%%%%%%%%%%%%%%%%%%%%%

% References are important to the reader; therefore, each citation must be complete and correct. If at all possible, references should be commonly available publications.

\bibliographystyle{ieeetr}
\bibliography{my_bib}

\end{document}